\documentclass[runningheads]{llncs}
\usepackage[T1]{fontenc}
\usepackage{graphicx}
\graphicspath{{media/}}     % organize your images and other figures under media/ folder
\usepackage{lipsum}
\usepackage{amsmath}
\usepackage{amsfonts}
\usepackage{algorithm}
\usepackage{array}
\usepackage[caption=false,font=normalsize,labelfont=sf,textfont=sf]{subfig}
\usepackage{textcomp}
\usepackage{stfloats}
\usepackage{url}
\usepackage{verbatim}
\usepackage{graphicx}
\usepackage{tablefootnote}
\usepackage[symbol]{footmisc}
\usepackage{hyperref}
\usepackage{mathtools, xspace, multirow, makecell, bbm, booktabs, algorithmicx}
\usepackage[group-separator={,}, group-minimum-digits={3}]{siunitx}
\usepackage[noend]{algpseudocode}
\makeatletter
\DeclareRobustCommand\onedot{\futurelet\@let@token\@onedot}
\def\@onedot{\ifx\@let@token.\else.\null\fi\xspace}

\def\eg{\emph{e.g}\onedot} 
\def\ie{\emph{i.e}\onedot} 
 
\def\etc{\emph{etc}\onedot}

\def\etal{\emph{et al}\onedot}
\makeatother

\newcommand{\INPUT}{\textbf{Input:}} % Use Input in the format of Algorithm
\newcommand{\OUTPUT}{\textbf{Output:}} % Use Output in the format of Algorithm

\usepackage{color}

\begin{document}
\title{Multi-Style Facial Sketch Synthesis through Masked Generative Modeling}
%
%\titlerunning{Abbreviated paper title}
% If the paper title is too long for the running head, you can set
% an abbreviated paper title here
%
\author{Bowen~Sun \and
Guo~Lu \and
Shibao~Zheng}
% \author{Bowen~Sun\inst{1}\orcidID{0000-1111-2222-3333} \and
% Second Author\inst{2,3}\orcidID{1111-2222-3333-4444} \and
% Shibao~Zheng\inst{3}\orcidID{2222--3333-4444-5555}}

%
% \authorrunning{BW~Sun et al.}
% First names are abbreviated in the running head.
% If there are more than two authors, 'et al.' is used.
%
\institute{Department of Electronic Engineering, Shanghai Jiao Tong University, Shanghai, China\\
% \email{lncs@springer.com}\\
% \url{http://www.springer.com/gp/computer-science/lncs} \and
% ABC Institute, Rupert-Karls-University Heidelberg, Heidelberg, Germany\\
\email{\{sunbowen,luguo2014,sbzh\}@sjtu.edu.cn}}
\maketitle              % typeset the header of the contribution
%
% \begin{abstract}
% The abstract should briefly summarize the contents of the paper in
% 150--250 words.

% \keywords{First keyword  \and Second keyword \and Another keyword.}
% \end{abstract}
\begin{abstract}
The facial sketch synthesis (FSS) model, capable of generating sketch portraits from given facial photographs, holds profound implications across multiple domains, encompassing cross-modal face recognition, entertainment, art, media, among others. However, the production of high-quality sketches remains a formidable task, primarily due to the challenges and flaws associated with three key factors: (1) the scarcity of artist-drawn data, (2) the constraints imposed by limited style types, and (3) the deficiencies of processing input information in existing models. To address these difficulties, we propose a lightweight end-to-end synthesis model that efficiently converts images to corresponding multi-stylized sketches, obviating the necessity for any supplementary inputs (\eg, 3D geometry). In this study, we overcome the issue of data insufficiency by incorporating semi-supervised learning into the training process. Additionally, we employ a feature extraction module and style embeddings to proficiently steer the generative transformer during the iterative prediction of masked image tokens, thus achieving a continuous stylized output that retains facial features accurately in sketches. The extensive experiments demonstrate that our method consistently outperforms previous algorithms across multiple benchmarks, exhibiting a discernible disparity.
% Face sketch synthesis (FSS) presents a promising solution to this challenge; however, existing FSS methods suffer from inherent flaws mainly including low quality and limited style diversity. 
\end{abstract}

% keywords can be removed
\section{Introduction}\label{sec:introduction}
The facial sketch synthesis (FSS) model~\cite{choiFacePhotoSketchSynthesis2023,fanFacialSketchSynthesisNew2022,gaoHumanInspiredFacialSketch2023a,tangFaceSketchSynthesis2003,zhuSketchTransformerNetworkFace2021}, which generates sketch portraits based on provided facial photographs, has made remarkable advancements across various domains. 
It not only offers a promising solution to the challenge of photo-sketch face recognition~\cite{tangFaceSketchSynthesis2003} in security fields but also serves as a pivotal component in content generation applications, encompassing entertainment, arts, media, and more. 
Consequently, FSS has garnered increasing attention, warranting our dedicated focus.

In the past few years, the significant progress made in FSS is largely attributed to the development of the generative adversarial network (GAN)~\cite{goodfellowGenerativeAdversarialNetworks2020}.
The GAN-based structures are the main generators for most of FSS methods.
To further enhance the performance of synthesized sketches, diverse techniques such as parametric sigmoid~\cite{zhangRobustFaceSketch2018}, semi-supervised learning~\cite{chenSemiSupervisedLearningFace2019}, 3D geometry~\cite{gaoHumanInspiredFacialSketch2023a}, semantic adaptive normalization~\cite{parkSemanticImageSynthesis2019a}, \etc, are also involved in the construction of models.
While these specifically designed methodologies, style transfer and stylization algorithms can be adapted to encompass FSS as a subsidiary application as well.
% Several generative models have been proposed for FSS, drawing inspiration from the generative adversarial network (GAN)~\cite{goodfellowGenerativeAdversarialNetworks2020} and self-attention mechanisms in transformers~\cite{vaswaniAttentionAllYou2017a}. 

Although numerous attempts have made by aforementioned models, it is regrettable that the pursuit of this task remains arduous, a consequence from the challenges mainly associated with the following significant reasons.
The foremost issue lies in the lack of hand-drawn photo-sketch pairs, which poses a challenge in meeting the training requirements of data-hungry deep neural networks, despite the early attention~\cite{choiDataInsufficiencySketch2012} given to this matter.
The acquisition of professional sketches incurs substantial costs, resulting in a limited number of samples even in the most recent and extensive publicly available database, FS2K~\cite{fanFacialSketchSynthesisNew2022}, which contains a mere \num[group-separator={,}]{2104} photo-sketch pairs.
Moreover, the practical styles of sketches are diverse, while training data usually provide limited types or even only one style, which obviously constrain the generative possibility of FSS models.
Furthermore, the practical styles of sketches exhibit significant diversity, whereas the available training data typically encompass limited or even solitary styles. 
This inherent constraint evidently impedes the generative potential of FSS models.
The third point pertains to the subpar quality of generation caused by inherent design flaws in the current algorithms, such as their inability to effectively handle interference from lighting and background, as well as their requirement for additional complex predictive information, such as depth or 3D geometry.
% Furthermore, the generation process of these methods necessitates additional predictions, such as depth and 3D information, which inevitably amplify both the complexity and inaccuracy of the model. 

\begin{figure}[t]
    \centering
    \includegraphics[width=\linewidth]{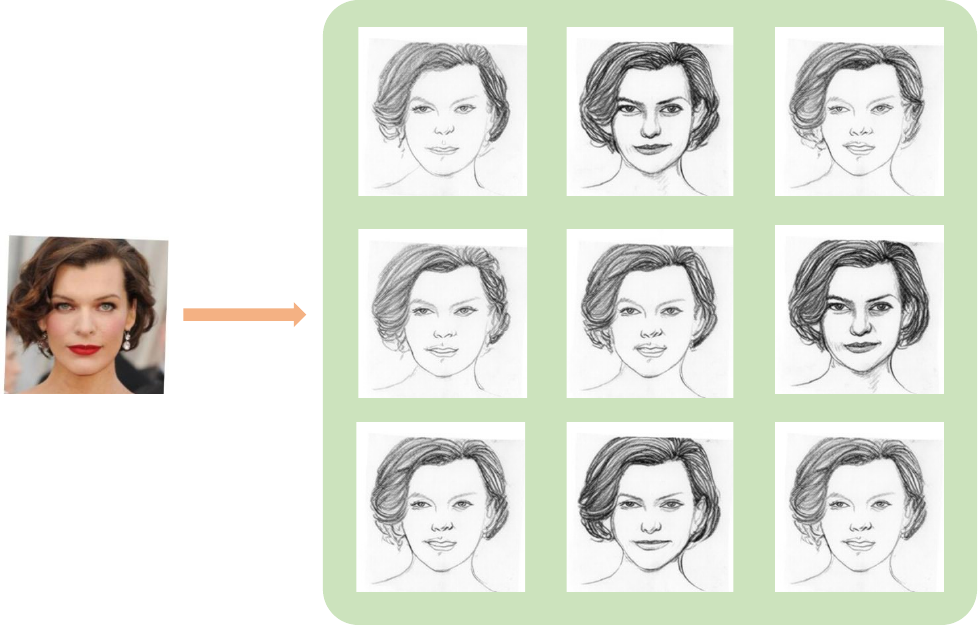}
    \caption{Illustration for multi-style FSS. By taking an facial image as the sole input, our method is able to stably generate corresponding sketch with multiple styles that are not contained in the training set.}
    \label{fig:illustration}
\end{figure}

% To address these challenges, we propose the algorithm that efficiently generates high-quality facial sketches with multiple styles. 
% We use a semi-supervision paradigm to break the constraint due to small scale dataset and condition embedding to realize multiple intermediate styles.
% To be specific, a masked generative transformer constitutes the base architecture of the model and is pre-trained with synthesized sketches before fine-tuned with real hand-drawn data. 
% The generation process is constrained in the token space of VQ-GAN~\cite{esserTamingTransformersHighresolution2021a} to decrease the computational consumption. 
% Thanks to the feature from a CLIP~\cite{radfordLearningTransferableVisual2021a} encoder, the generative transformer is finally able to reconstruct the quantified sketch tokens for the decoder.
% During training, the mechanism of masked image model (MIM)~\cite{changMaskgitMaskedGenerative2022} is adopted to the tokens, which is expected to be recovered by the transformer model.

To tackle these challenges, we propose an algorithm that efficiently generates high-quality facial sketches encompassing multiple styles, as shown in Fig.\ref{fig:illustration}. 
Our approach employs a semi-supervised paradigm to overcome the limitations imposed by a small-scale dataset, while leveraging condition embedding to enable the realization of various intermediate styles.
Specifically, our model is built upon a masked generative transformer~\cite{changMaskgitMaskedGenerative2022}, serving as the foundational architecture. 
Initially, the transformer is pre-trained using synthesized sketches and subsequently fine-tuned using real hand-drawn data. 
To optimize computational efficiency, we constrain the generation process within the token space of VQ-GAN~\cite{esserTamingTransformersHighresolution2021a}.
By incorporating the feature obtained from an encoder, our generative transformer successfully reconstructs quantified sketch tokens for the decoder. 
During the training process, we employ the masked image modeling mechanism on the tokens, with the expectation that they will be recovered by the transformer model.
Our contributions can be summarized as follows:
\begin{itemize}
    \item The proposed method successfully achieves stable multi-style FSS by utilizing a 2D photograph as the exclusive input without any dependence on data from other modalities, in contrast to some prior approaches.
    \item We employ the masked generative modeling and semi-supervised learning to address the challenge of limited availability of high-quality datasets containing face-sketch pairs.
    \item The impartial experiments have demonstrated that our algorithm consistently and significantly outperforms the current algorithms, both in terms of quantitative generation quality and structural similarity, as well as in qualitative visual comparison.
\end{itemize}
\section{Related Works}\label{sec:related_works}
In this section, we provide a comprehensive review of previous FSS methods, highlighting the techniques that have inspired our proposed approach.

\textbf{FSS}. The introduction of learning schemes has led to significant advancements in early traditional FSS that rely on heuristic image transformations. 
This progress has given rise to the development of these models, including Bayesian inference models~\cite{chenExamplebasedFacialSketch2001,pengMultipleRepresentationsbasedFace2015,zhouMarkovWeightFields2012}, representation learning models~\cite{jiLocalRegressionModel2011,zhangFaceSketchphotoSynthesis2011}, and subspace learning models~\cite{huangCoupledDictionaryFeature2013,liuNonlinearApproachFace2005}.
In recent times, the burgeoning advancements in content generation by artificial intelligence have significantly enhanced the field of deep facial sketch synthesis, particularly through the utilization of style transfer~\cite{jingNeuralStyleTransfer2020,zhuUnpairedImagetoimageTranslation2017} and image translation~\cite{saxenaComparisonAnalysisImagetoImage2022} methodologies.
For instance, the Pix2Pix~\cite{isolaImagetoimageTranslationConditional2017} model facilitates the transformation of source images into target images by replacing the generator of original conditional GAN with U-Net~\cite{ronnebergerUNetConvolutionalNetworks2015a}. 
Subsequently, this approach has been enhanced to address its inherent instability by incorporating the feature matching loss, as proposed by Wang \etal~\cite{wangHighresolutionImageSynthesis2018}, which also enables the generation of high-resolution images with dimensions of 2048$\times$1024.
Through the utilization of spatially-adaptive normalization, also known as SPADE~\cite{parkSemanticImageSynthesis2019a}, and its subsequent advancement, DySPADE~\cite{gaoHumanInspiredFacialSketch2023a}, the retention of semantic information is augmented, consequently enhancing the performance of sketch generation.
In order to train the FSS models, numerous researchers have collected datasets~\cite{fanFacialSketchSynthesisNew2022,wangFacePhotosketchSynthesis2008,zhangCoupledInformationtheoreticEncoding2011} containing hand-drawn facial sketches. 
Nevertheless, the scarcity of comprehensive datasets of this nature hitherto persists owing to the exorbitant expenses involved in their creation, which necessitates the acquisition of paired photographs and corresponding sketches.

\textbf{Masked Generative Transformers}. With the widespread adoption of transformers~\cite{vaswaniAttentionAllYou2017a} in the field of computer vision, numerous network architectures~\cite{hoogeboomSimpleDiffusionEndtoend2023,peeblesScalableDiffusionModels2023} and algorithms have been developed, significantly enhancing the quality of content generation based on diffusion models. 
Despite the superior training stability of diffusion models compared to GANs, they suffer from the drawback of high computational requirements. 
Drawing inspiration from MIM~\cite{heMaskedAutoencodersAre2022}, previous studies have introduced masked generative transformers~\cite{changMaskgitMaskedGenerative2022,changMuseTextToImageGeneration2023a,patilAMUSEdOpenMUSE2024a} as a viable solution to tackle this concern. 
These self-supervised approaches aim to recover masked image tokens using a transformer model, thereby enabling the reconstruction of the original images. 
Consequently, Muse~\cite{changMuseTextToImageGeneration2023a} accomplishes efficient text-to-image generation while exhibiting zero-shot editing capabilities.
Based on this framework, we propose a generative model aimed at constructing facial sketches from completely masked tokens, with the guidance provided by the facial features extracted from photographs.

% CLIP~\cite{radfordLearningTransferableVisual2021a}
\section{Method}\label{sec:method}
We commence by introducing the problem formulation in Section~\ref{subsec:problem_formulation} and the corresponding loss functions in Section~\ref{subsec:loss}. 
Subsequently, we provide the details of the model's architecture and the comprehensive training procedure in Section~\ref{subsec:architecture}.

\begin{figure}[t]
    \centering
    \includegraphics[width=\linewidth]{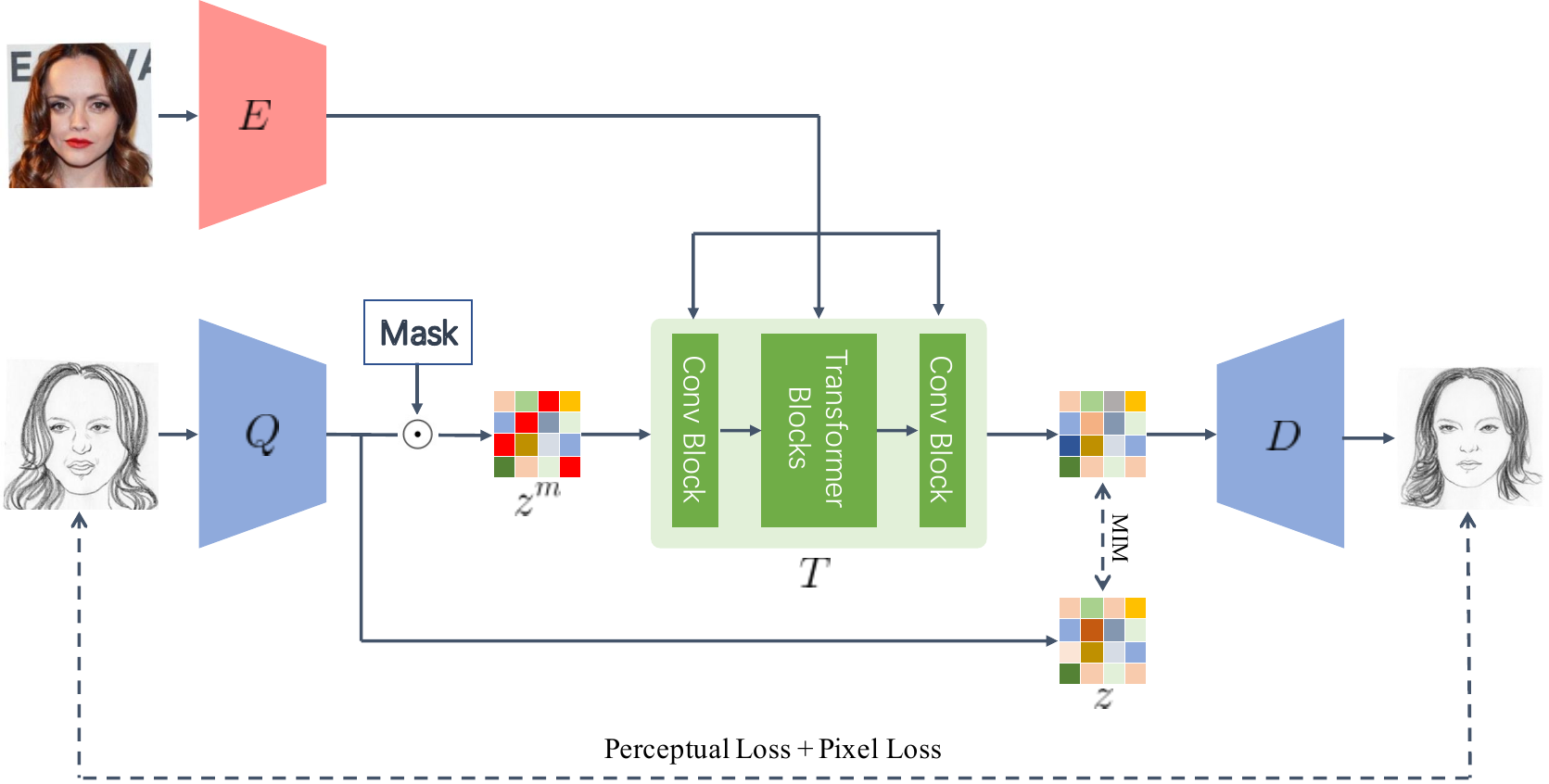}
    \caption{The architecture and training pipeline for our method. The red portion in $z^m$ represents the masked tokens.}
    \label{fig:architecture}
\end{figure}

\subsection{Problem Formulation}\label{subsec:problem_formulation}
Taking the facial photograph $x\in\mathcal{X}$ and style condition $s\in\mathcal{S}$ as inputs, our objective is to discover a mapping $F:\mathcal{X} \times \mathcal{S} \rightarrow \mathcal{Y}$, where $y\in\mathcal{Y}$ represents the identity-invariant sketch subject to the desired style $s$.
To accomplish this target, the complete model comprises the VQ-tokenizer $Q$ (used exclusively during training), the feature encoder $E$, the transformer $T$, and the decoder $D$, forming the pipeline as illustrated in Fig.~\ref{fig:architecture}.

The predominant manipulations occur within the latent space, where the transformer model recovers masked image tokens.
If sequence $\{z_t\}$ represents the true latent tokens gradually masked with respect to discrete time step $t=0,1,...,\mathcal{T}$, the transformer $T$ is trained to iterativelly reconstruct the initial $z_0$ from $z_\mathcal{T}$.
Mathematically, given the totally masked latent token $z_\mathcal{T}$, the prediction $\Tilde{z}_{t}$ from the transformer is obtained by
\begin{equation}\label{eq:iteration}
\left\{
    \begin{array}{l}
    \Tilde{z}_{t-1} = \boldsymbol{q}(T(E (x), s, \Tilde{z}_{t})), 1 \leq t \leq \mathcal{T}, \\
    \Tilde{z}_\mathcal{T} = z_\mathcal{T}, \\
    \end{array}
\right.
\end{equation}
where $\boldsymbol{q}$ corresponds to the element-wise quantization proposed in VQ-GAN~\cite{esserTamingTransformersHighresolution2021a}.
To be specific, the term $E(x)$ in Eq.~\eqref{eq:iteration} encompasses the hidden states of the encoder $E$, rather than solely representing the features output by its final layer.
Finally, the synthesized sketch $\Tilde{y}_0$ in inference process is obtained by 
\begin{equation}\label{eq:problem_formulation}
    \Tilde{y}_0 = D (\boldsymbol{q} (\Tilde{z}_0)).
\end{equation}
Based on Eq.~\eqref{eq:problem_formulation}, we can find that the quality of the synthesized image relies on two crucial factors: the decoder itself and the disparity between $z_0$ and $\Tilde{z}_0$.

In pursuit of this objective, we have devised a training procedure aimed at optimizing $T_\theta$ and $D_\theta$, whose parameters are denoted by $\theta$.
Given the paired photos $x$ and sketches $y$ in the training dataset, the latent image tokens $z$ are obtained by applying the function $z=Q(y)$. 
In contrast to inference, our training process involves a single step sampling of the mask for each image, followed by the prediction using $\Tilde{z}=\boldsymbol{q}(T_\theta(E(x), s, z^m))$, where $z^m$ represents the masked tokens. 
The synthesized sketch $\Tilde{y}$ is subsequently obtained through $D_\theta(\Tilde{z})$.
Naturally, our objective function aims to optimize
\begin{equation}
    \mathop{\arg\min}\limits_{\theta} \mathbb{E}_{x, y} \left[\mathcal{D}_1(\Tilde{z}, z)+\mathcal{D}_2(\Tilde{y}, y) \right],
\end{equation}
where $\mathcal{D}_1$ and $\mathcal{D}_2$ are two functions measuring distance.
Their specific forms depend on the subsequently introduced loss functions in Section~\ref{subsec:loss} for training $T_\theta$ and $D_\theta$.

\subsection{Loss Functions}\label{subsec:loss}
Drawing upon the architecture illustrated in Fig.~\ref{fig:architecture}, we employ a two-stage training procedure to optimize the transformer $T$ and decoder $D$ individually. 
To be specific, MIM loss serves as the sole loss function employed during training of $T$, while the others are used for taining $D$.
Concurrently, we leverage a pre-trained feature encoder $E$ and VQ-tokenizer $Q$, whose parameters remain fixed throughout the entire training process.

\textbf{MIM Loss}. In VQ-GAN, the latent tokens $z$ with a dimension of $N$ is constructed using $N$ discrete labels, derived from a codebook comprising $C$ distinct classes.
Subsequently, we sample masked tokens $z^m$ using a cosine mask schedule that masks a fraction of $z$ with a masking ratio $R=\cos(\pi \cdot r/2)$, where $r$ is randomly drawn from a uniform distribution $U(0,1)$.
For latent tokens $z$ of length $N$, the mask $\{m_i\}_{i=1}^N$ represents a binary sequence associated with $z$. 
The masked tokens $z^m$ is derived by substituting the token at position $i$ of $z$ with a designated symbol ($-100$ in this work) when $m_i=1$, \ie, the $i$-th position of $z^m$ (denoted by $z^m_i$) is obtained by
\begin{equation}\label{eq:z_m}
   z^m_i = 
   \left\{
    \begin{array}{l}
    z_i, m_i=0, \\
    -100,m_i=1.
    \end{array}
\right.
\end{equation}
As the primary objective of the generative transformer model $T_\theta$ is to reconstruct the masked portion within $z^m$, the MIM loss function $L_{MIM}$ is defined as
\begin{equation}\label{eq:l_mim}
    L_{MIM} = -\frac{1}{N-M} \sum_{i=1}^N m_i \left[ \sum_{c=1}^C  l_{ic} \log (T_\theta(E(x), s, z^m)) \right],
\end{equation}
where $M=\sum_{i=1}^N m_i$, and $l_{ic}$ denotes the one-hot ground truth label of the $i$-th position in $z$.
In Eq.~\eqref{eq:l_mim}, the utilization of the negative log-likelihood reveals that $L_{MIM}$ is a cross-entropy loss function specifically applied to masked tokens.

\textbf{Pixel Loss}. We deploy pixel loss function to measure the pixel-wise distance between synthesized and target sketch image.
If we denote $\|\cdot\|_1$ as the $\ell_1$-norm, the pixel loss $L_{pix}$ is defined by
\begin{equation}\label{eq:l_pix}
    L_{pix} = \| \Tilde{y} - y \|_1.
\end{equation}

\textbf{Perceptual Loss}. In order to assure the feature similarity between synthesized sketch and the ground truth, we additionally employ the perceptual loss:
\begin{equation}\label{eq:L_pcpt}
    L_{pcpt} = \sum_l W_l\| f_l(\Tilde{y}) - f_l(y) \|^2_2,
\end{equation}
where $W_l$ and $f_l$ are the $l$-th weight and layer's output, respectively.

\textbf{Objective}. In this work, the distance $\mathcal{D}_1(\Tilde{z}, z)$ is measured by $L_{MIM}$, while the $\mathcal{D}_2(\Tilde{y}, y)$ is decided by $L_{pix} $ and $ L_{pcpt}$ in practice, \ie,
\begin{equation}\label{eq:total_d}
\left\{
    \begin{array}{l}
    \mathcal{D}_1(\Tilde{z}, z)=L_{MIM}, \\
    \mathcal{D}_2(\Tilde{y}, y)=\lambda_1 L_{pix} + \lambda_2 L_{pcpt},
    \end{array}
\right.
\end{equation}
where $\lambda_1$ and $\lambda_2$ are weight parameters.

\begin{algorithm}[t]
\caption{Training Procedure}
\label{alg:train}
\INPUT Facial photo-sketch pair $(x,y)$; style parameter $s$; pre-trained encoder $E$, VQ-tokenizer $Q$ and decoder $D$;\\
\OUTPUT Well-trained network $T$ and finetuned $D$
    \begin{algorithmic}[1]
        \While {not converged}
        \State $z \gets Q(y)$, $z^m \gets $ Eq.~\eqref{eq:z_m}
        \State $L_{MIM} \gets $ Eq.~\eqref{eq:l_mim}
        \State Update parameters of $T$ to reduce $L_{MIM}$
        \EndWhile

        \While {not converged}
        \State $z \gets Q(y)$, $z^m \gets $ Eq.~\eqref{eq:z_m}
        \State $\Tilde{z} \gets \boldsymbol{q}(T_\theta(E(x), s, z^m))$, $\Tilde{y} \gets D(\Tilde{z})$ 
        \State $L_{MIM} \gets $ Eq.~\eqref{eq:l_mim}, $L_{pix} \gets $ Eq.~\eqref{eq:l_pix}, $L_{pcpt} \gets $ Eq.~\eqref{eq:L_pcpt}
        \State Update parameters of $T$ and $D$ to reduce $\mathcal{D}_1+\mathcal{D}_2$ in Eq.~\eqref{eq:total_d}
        \EndWhile
    \end{algorithmic}
\end{algorithm}

\subsection{Architecture and Algorithm}\label{subsec:architecture}
We hereby present a comprehensive exposition of all components within our framework, as well as an elaborate description of the procedure employed in our training algorithm.

\textbf{Feature Encoder}. The encoder $E$ plays a crucial role in acquiring feature embeddings within the latent space, thereby guiding the generation direction of the transformer model. 
To ensure optimal performance, we employ CLIP-L/14~\cite{radfordLearningTransferableVisual2021a} as our chosen encoder, as it has been reported to possess exceptional face recognition capabilities~\cite{bhatFaceRecognitionAge2023}.
The intermediate and final hidden states are injected to the transformer model by cross-attention mechanism and adaptive normalization layers~\cite{perezFilmVisualReasoning2018}, respectively.

\textbf{Transformer}. We adopt a modified version of U-ViT~\cite{hoogeboomSimpleDiffusionEndtoend2023} architecture, sharing the same structure applied by~\cite{patilAMUSEdOpenMUSE2024a}, as our transformer model $T$.
Our model outputs tokens of compact dimensions, specifically 16$\times$16, eliminating the necessity for downsampling and upsampling blocks.
The model $T$ receives a sequence of masked image token with the same dimension, \ie, 256 (16$\times$16) and facial photo feature from $E$ with dimension of \num[group-separator={,}]{1024}.
The model $T$ is fed a sequence of masked image tokens, each having the same dimension of its output, along with facial photo features obtained from $E$, which possess a dimension of \num[group-separator={,}]{1024}.

\textbf{VQ-Tokenizer/Decoder}. To effectively achieve the generation process, we perform manipulation within the compressed latent space facilitated by VQ-GAN~\cite{esserTamingTransformersHighresolution2021a}. 
The model comprises a tokenizer $Q$, which downsample images into the latent space, and a decoder $D$, which upsamples these tokens to generate images with a resolution of 256$\times$256.
By converting the image to tokens with size of 16$\times$16, the model effectively reduces the dimension for transformer $T$ to process.
Remarkably, this model does not incorporate any self-attention layers and employs a vocabulary size of \num[group-separator={,}]{8192}.

\textbf{Algorithm}. Utilizing the aforementioned architectures and loss functions, we proceed with training the model in a two-stage framework, as shown in the table of Algorithm~\ref{alg:train}. 
In the initial stage (step 1-4), namely the pre-training, our focus lies solely on training the transformer model $T$ using the MIM loss. 
Following the sampling of $x$, $y$, and $r$, we employ the VQ-tokenizer $Q$ to obtain $z$, subsequently subjecting it to a random mask operation, thereby transforming it into $z^m$.
The parameters of $T_\theta$ are adjusted to minimize the distance $\mathcal{D}_1(\Tilde{z}, z)$, where $\Tilde{z}=T_\theta(E(x),s,z^m)$.
In the second stage of fine-tuning (step 5-9), the decoder $D$ generates the synthesized output $\Tilde{y}$ by utilizing $\Tilde{z}$ as its input.
Subsequently, we optimize the perceptual loss and pixel loss to further refine the performance of $D$.

\section{Experiments}\label{sec:experiments}
We provide the comprehensive experimental results in this section.
Comparisons with various methods and the ablation study are conducted to analyze the performance.

\begin{figure}[t]
    \centering
    \includegraphics[width=\linewidth]{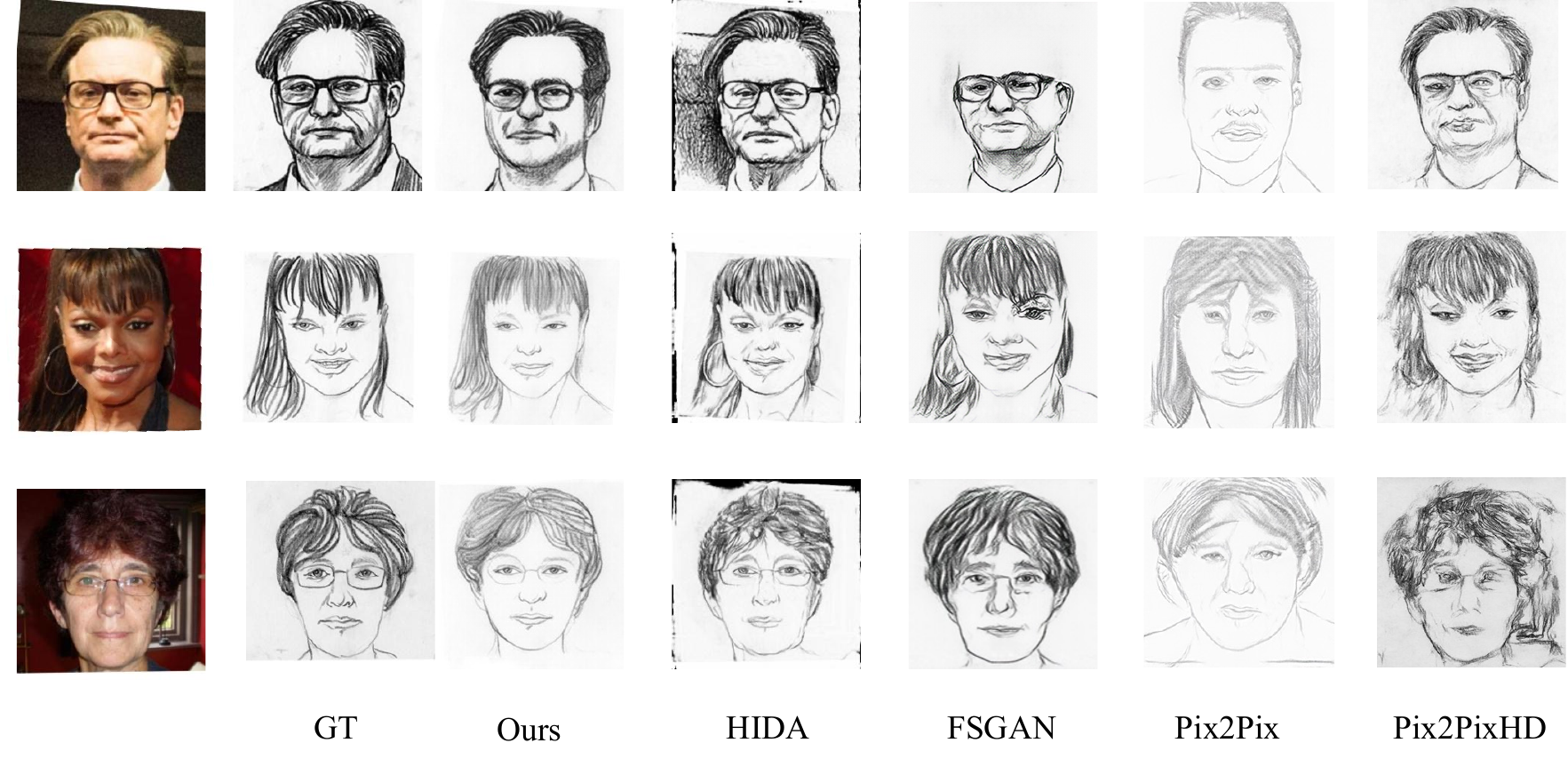}
    \caption{Comparison of sketches synthesized by various typical methods.}
    \label{fig:exp}
\end{figure}

\subsection{Experimental Settings}
\textbf{Datasets}. Our model undergoes pre-training using the CelebA dataset~\cite{liuDeepLearningFace2015a}, followed by fine-tuning using the FS2K database~\cite{fanFacialSketchSynthesisNew2022}.
The CelebA dataset comprises a total of \num[group-separator={,}]{202599} facial images exhibiting a wide range of poses and backgrounds.
We generate the corresponding sketches of images in CelebA by \cite{gaoHumanInspiredFacialSketch2023a} and eliminate the background in sketches by parsing the facial area.
Contrasting with the larger scale of the CelebA dataset, the FS2K database comprises a relatively smaller collection of \num[group-separator={,}]{2104} image-sketch pairs, which encompass three distinct sketch styles, as well as a diverse range of image backgrounds and lighting conditions.
Despite its smaller size, the FS2K stands out due to its high-quality hand-drawn facial sketches that are meticulously paired with corresponding photographs. 
Consequently, it is selected for the fine-tuning process, contributing to the overall quality and effectiveness of the whole model.

\textbf{Parametric Settings}. In our study, we assign the hyper-parameters $\lambda_1$ and $\lambda_2$ the values of 4 and 10, respectively. 
The training process takes place on a server consisting of four RTX3090 GPUs, with a batch size of 4 for each GPU and an accumulation step of 8. 
The parameters of the feature encoder $D$ and VQ-tokenizer $Q$ remain frozen throughout all training procedures. 
In the pre-training phase, the transformer $T$ undergoes training with learning rate of $3\times 10^{-4}$ for a total of \num[group-separator={,}]{10000} steps.
We subsequently finetune the decoder together with the transformer model for another \num[group-separator={,}]{4000} steps.

\textbf{Criteria}. To comprehensively evaluate the performance of the compared methods, we employ 5 indices as evaluation criteria, namely the Learned Perceptual Image Patch Similarity (LPIPS)~\cite{zhangUnreasonableEffectivenessDeep2018} Fr\'{e}chet Inception distance (FID)~\cite{heuselGansTrainedTwo2017}, Structural Similarity (SSIM)~\cite{wangImageQualityAssessment2004}, Feature Similarity Measure (FSIM)~\cite{zhangFSIMFeatureSimilarity2011}, and Structure Co-Occurrence Texture (SCOOT)~\cite{fanScootPerceptualMetric2019}.
These metrics are commonly applied for assessing the structural, distributional, and perceptual discrepancies between synthesized and ground truth images in the test set.
Significantly, the attainment of lower LPIPS and FID scores or higher SSIM, FSIM, and SCOOT scores denotes superior performance, which is indicated by symbols $\downarrow$ and $\uparrow$.

\begin{table}[t]
\setlength{\tabcolsep}{8pt}
    \caption{Quality of the generation across various benchmarks.}\label{tab:scores}
    \centering
    \begin{tabular}{l|ccccc}
    \toprule
            & LPIPS$\downarrow$ & FID$\downarrow$ & SSIM$\uparrow$ & FSIM$\uparrow$ & SCOOT$\uparrow$ \\
    \midrule
    Pix2Pix   &0.474  &40.98  &0.890  &0.288  &0.413 \\
    Pix2PixHD &0.456  &45.27  &0.871  &0.301  &0.455 \\
    CycleGAN  &0.477  &29.13  &0.833  &0.298  &0.337 \\
    MDAL      &0.482  &62.58  &0.865  &0.302  &0.419 \\
    CartoonGAN&0.522  &74.30  &0.859  &0.297  &0.372 \\
    % GENRE     &  &  &  &  & \\
    FSGAN     &0.439  &42.47  &0.869  &0.306  &0.467 \\
    HIDA      &0.403  &51.84  &0.883  &0.363  &0.528 \\
    Ours      &\textbf{0.337}  &\textbf{26.79}  &\textbf{0.909}  &\textbf{0.377}  &\textbf{0.533} \\
    \bottomrule
    \end{tabular}
\end{table}

\subsection{Evaluation on Quality}
In this section, we undertake a comparative analysis on the qulity of synthesized sketchs between our proposed method and a range of state-of-the-art FSS techniques, namely HIDA~\cite{gaoHumanInspiredFacialSketch2023a}, FSGAN~\cite{fanFacialSketchSynthesisNew2022}, 
% GENRE~\cite{liHighqualityFaceSketch2021}, 
% SCA-GAN~\cite{yuRealisticFacePhoto2020}, 
CartoonGAN~\cite{chenCartoonGANGenerativeAdversarial2018}, MDAL~\cite{zhangFaceSketchSynthesis2018}, CycleGAN~\cite{zhuUnpairedImagetoimageTranslation2017}, Pix2Pix~\cite{isolaImagetoimageTranslationConditional2017}, and Pix2PixHD~\cite{wangHighresolutionImageSynthesis2018}. 
A portion of the synthesis results has been sourced from~\cite{fanFacialSketchSynthesisNew2022}, while all the results have been fairly evaluated using identical criteria and settings.

In Fig.~\ref{fig:exp}, we present a subset of results generated by several state-of-the-art methods on FS2K~\cite{fanFacialSketchSynthesisNew2022} for the purpose of comparing and analyzing their performance.
Our method demonstrates evident style consistency with the ground truth sketches and exhibits a high degree of feature similarity with the input photograph.
Our algorithm possesses the capability to depict portraits with clear and coherent lines, capturing intricate details such as hairstyles, eyes, noses, mouths, and other facial features.
When compared to other algorithms, including the latest HIDA~\cite{gaoHumanInspiredFacialSketch2023a}, our method excels in accurately separating the subject from the background and preserving crucial details, even in scenarios where the contrast between the subject and the background is not pronounced (first row in Fig.~\ref{fig:exp}).
Moreover, the outputs generated by HIDA demonstrate irregular black borders, thereby significantly compromising the overall image quality. 
Additionally, our approach surpasses Pix2Pix in terms of feature similarity between synthesized sketches and the original input image; and it outperforms Pix2PixHD and FSGAN in terms of clarity.

To conduct a comprehensive quantitative analysis on the generation quality of each algorithm, we present the scores of multiple metrics in Tab. \ref{tab:scores}. 
Consequently, our method surpasses all other state-of-the-art methods across all benchmarks. 
Our algorithm demonstrates the lowest losses in terms of LPIPS and FID metrics, signifying its exceptional visual perception similarity and data distribution resemblance compared to the target within the test set. 
Furthermore, we have achieved the highest scores in terms of SSIM, FSIM, and SCOOT metrics, indicating a robust structural similarity between the generated face sketches and the target.

\subsection{Multi-Style Output}
While achieving high-quality generation, our algorithm also accomplishes style interpolation, enabling it to generate intermediate styles beyond the limited set of styles in the training dataset, thus achieving a more diverse and rich style output.

\begin{figure}[t]
    \centering
    \includegraphics[width=\linewidth]{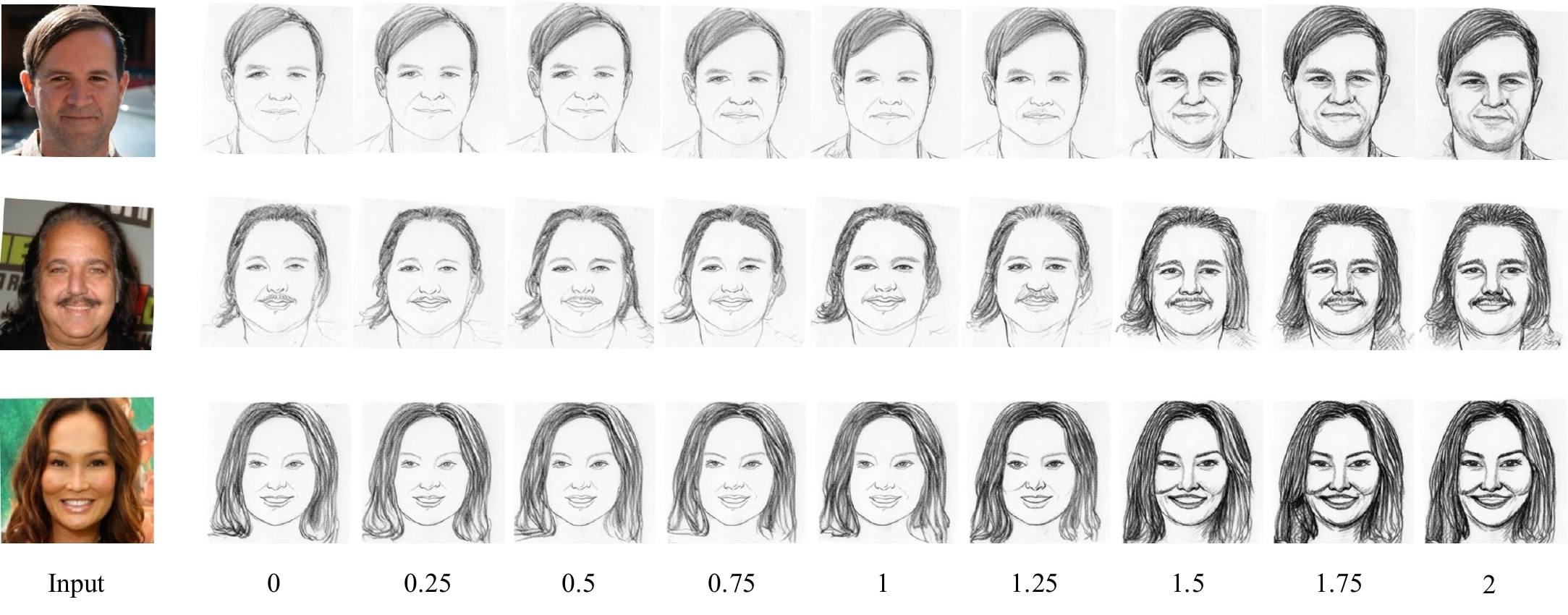}
    \caption{Synthesized sketches with different style parameters. The number at the bottom of each column is the style parameter $s$.}
    \label{fig:style_inter}
\end{figure}

% 如Fig.~\ref{fig:style_inter}所示，通过调整风格参数$s$，我们可以得到一系列中间风格。不同于当前的大多数人脸草图生成算法，例如HIDA，依赖于one-hot标签从而仅能生成训练集中所包含的特定几种风格，我们的算法将其视为一个参数。在推理过程中，只要调整参数范围就能生成一串渐变的草图风格。这大大扩展了模型输出风格的多样性，并且有助于其他领域的应用，例如在跨模态人脸草图识别领域的数据扩增等。

As shown in Fig.~\ref{fig:style_inter}, by adjusting the style parameter $s$, a series of intermediate styles can be obtained. In contrast to most existing face sketch generation algorithms, such as HIDA, which rely on one-hot labels and can only generate specific styles contained in the training set, our algorithm treats the style as a parameter. During the inference process, a continuous range of sketch styles can be generated simply by adjusting the parameter. This significantly expands the diversity of model output styles and contributes to applications in other domains, such as data augmentation in cross-modal face sketch recognition.

\section{Conclusion}
% 为了应对当前人脸草图生成算法开发中遇到的数据不足，风格有限，输入复杂等问题，本文提出了一种创新性的基于masked generative modeling的高效人脸草图生成算法。该算法通过半监督和自监督学习，不仅缓解了数据不足带来的挑战，而且其迭代恢复图像的方式避免了使用训练不稳定的GAN，GAN是当前主流草图人脸生成的基础模型。此外，我们的算法还具有当前大多数人脸草图生成算法所不具备的风格差值能力。这使得生成的草图不拘泥于训练集中的有限风格，而是能够稳定地无级生成介于不同风格间的中间风格。对于输入，我们的算法无需复杂的额外信息，仅需一张人脸照片即可生成对应的草图。一系列公正的实验也证实了我们的算法具有更高的生成质量，在背景与前景分离和多风格生成等方面也更具优势。

To address the challenges of insufficient data, limited style, and complex inputs encountered in the development of current facial sketch generation algorithms, this paper proposes an innovative and efficient facial sketch generation algorithm based on masked generative modeling. This algorithm utilizes semi-supervised and self-supervised learning techniques, which not only alleviate the challenges caused by insufficient data but also avoid the use of training-unstable GANs, which are the fundamental models in the current mainstream sketch facial generation. Furthermore, our algorithm possesses the ability to interpolate between different styles, a feature lacking in most existing facial sketch generation algorithms. This enables the generated sketches to transcend the limited styles present in the training set and consistently generate intermediate styles between different ones. Our algorithm requires no complex additional information for input; a single facial photograph is sufficient to generate the corresponding sketch. A series of fair experiments also confirm the higher generation quality of our algorithm, as well as its advantages in background and foreground separation and multi-style generation.
%
% ---- Bibliography ----
%
% BibTeX users should specify bibliography style 'splncs04'.
% References will then be sorted and formatted in the correct style.
%
\bibliographystyle{splncs04}
\bibliography{reference}
\end{document}